\documentclass[letterpaper, 10 pt, journal, twoside]{IEEEtran}

\IEEEoverridecommandlockouts                              

\usepackage{amsfonts}
\usepackage{amsmath}
\usepackage{amssymb}
\usepackage{algorithm}
\usepackage[noend]{algorithmic}
\usepackage{url}
\usepackage{orcidlink}
\usepackage[caption=false,font=footnotesize]{subfig}
\usepackage{hyperref}

\usepackage{fancyhdr}
\usepackage{xcolor}

\newtheorem{dfn}{Definition}
\newtheorem{prb}{Problem}

\newcommand{\blue}[1]{\textcolor{blue}{#1}}

\title{Making a Complete Mess and Getting Away with it: Traveling Salesperson Problems with Circle Placement Variants}

\author{David Woller\textsuperscript{\orcidlink{0000-0001-8809-3587}}, 
Masoumeh Mansouri\textsuperscript{\orcidlink{0000-0002-0997-5889}} and 
Miroslav Kulich\textsuperscript{\orcidlink{0000-0002-4527-7586}}
\thanks{Manuscript received: April, 2, 2024; Revised July, 8, 2024; Accepted August, 12, 2024.}
\thanks{This paper was recommended for publication by Editor Hanna Kurniawati upon evaluation of the Associate Editor and Reviewers' comments.}
\thanks{The work of David Woller and Miroslav Kulich was funded by the European Union under the project ROBOPROX (reg. no. CZ.02.01.01/00/22\_008/0004590).
The work of D. Woller was also funded by the Grant Agency of the Czech Technical University in Prague, under Grant SGS23/122/OHK3/2T/13.
Masoumeh Mansouri is a UK participant in Horizon Europe Project CONVINCE, and supported by UKRI grant number 10042096. For the purpose of open access, the author has applied a Creative Commons Attribution (CC BY) license to any Accepted Manuscript version arising.} 
\thanks{D. Woller and M. Kulich are with the Czech Institute of Informatics, Robotics, and Cybernetics, Czech Technical University in Prague, Czech Republic ({\tt\footnotesize wolledav@cvut.cz}, {\tt\footnotesize kulich@cvut.cz}).}%
\thanks{M. Mansouri is with the School of Computer Science, University of Birmingham, United Kingdom ({\tt\footnotesize m.mansouri@bham.ac.uk}).}%
\thanks{Digital Object Identifier (DOI): \url{https://doi.org/10.1109/LRA.2024.3445817}}
}

\fancypagestyle{FirstPage}{
\lhead{
\scriptsize
IEEE ROBOTICS AND AUTOMATION LETTERS. PREPRINT VERSION. ACCEPTED AUGUST, 2024 \\
© 2024 IEEE.  Personal use of this material is permitted.  Permission from IEEE must be obtained for all other uses, in any current or future media, including reprinting/republishing this material for advertising or promotional purposes, creating new collective works, for resale or redistribution to servers or lists, or reuse of any copyrighted component of this work in other works.}
}
\begin{document}

\maketitle

\markboth{IEEE Robotics and Automation Letters. Preprint Version. Accepted August, 2024}
{Woller \MakeLowercase{\textit{et al.}}: Traveling Salesperson Problems with Circle Placement Variants}

\begin{abstract}

This paper explores a variation of the Traveling Salesperson Problem, where the agent places a circular obstacle next to each node once it visits it.
Referred to as the Traveling Salesperson Problem with Circle Placement (TSP-CP), the aim is to maximize the obstacle radius for which a valid closed tour exists and then minimize the tour cost.
The TSP-CP finds relevance in various real-world applications, such as harvesting, quarrying, and open-pit mining. We propose several novel solvers to address the TSP-CP, its variant tailored for Dubins vehicles, and a crucial subproblem known as the Traveling Salesperson Problem on self-deleting graphs (TSP-SD). Our extensive experimental results show that the proposed solvers outperform the current state-of-the-art on related problems in solution quality.
\end{abstract}


\section{Introduction}
\thispagestyle{FirstPage}

\IEEEPARstart{T}{his} paper addresses a variation of the Traveling Salesperson Problem (TSP), in which the agent must visit each node in a weighted graph exactly once and place a circular obstacle
next to it. This problem is referred to as the Traveling Salesperson Problem with Circle Placement (TSP-CP). 
Similar to the TSP, the objective in the TSP-CP is to minimize the tour cost. However, TSP-CP has a different primary optimization criterion that involves maximizing the radius of the circular obstacle associated with each visited node. The TSP-CP is relevant to many real-world applications, such as autonomous harvesting~\cite{ullrich2014sugarbeet}, quarrying, and open-pit mining~\cite{mansouri2017multi}, to name only a few.

\begin{figure}[t]
    \centering
    \includegraphics[width=\columnwidth]{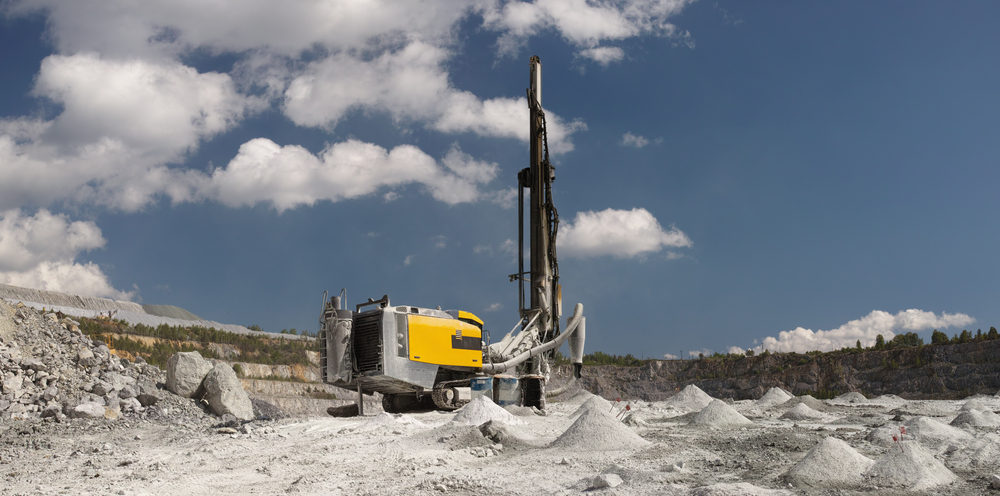}
    \caption{Blast hole drilling~\cite{Salienko20}.}
    \label{fig:drill_rig}
\end{figure}

\begin{figure*}[t!]
\centering
\subfloat[TSP-CP solution]{\includegraphics[width=0.34\textwidth]{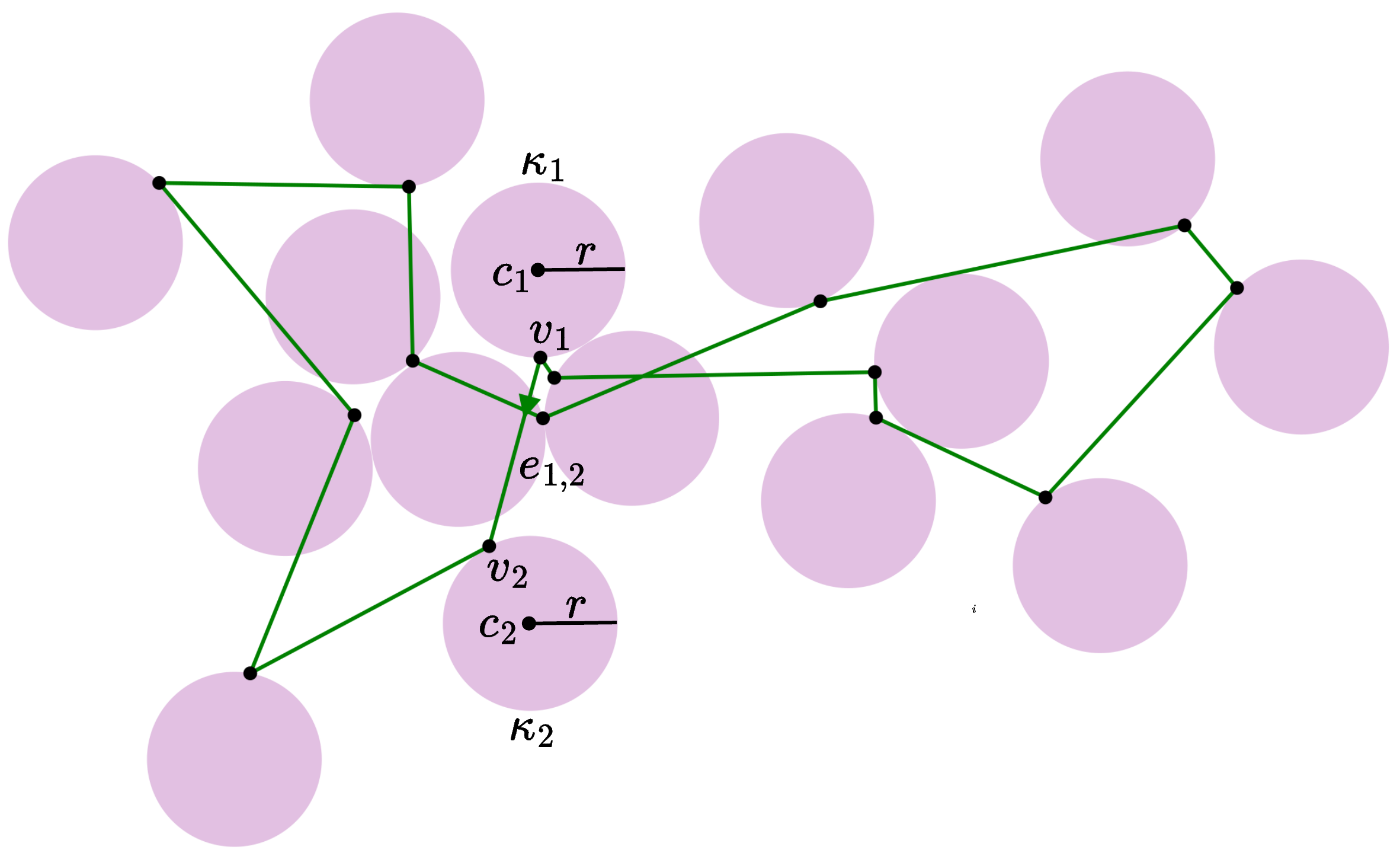}%
\label{fig:TSPCP}}
\hfil
\subfloat[DTSP-CP solution]{\includegraphics[width=0.34\textwidth]{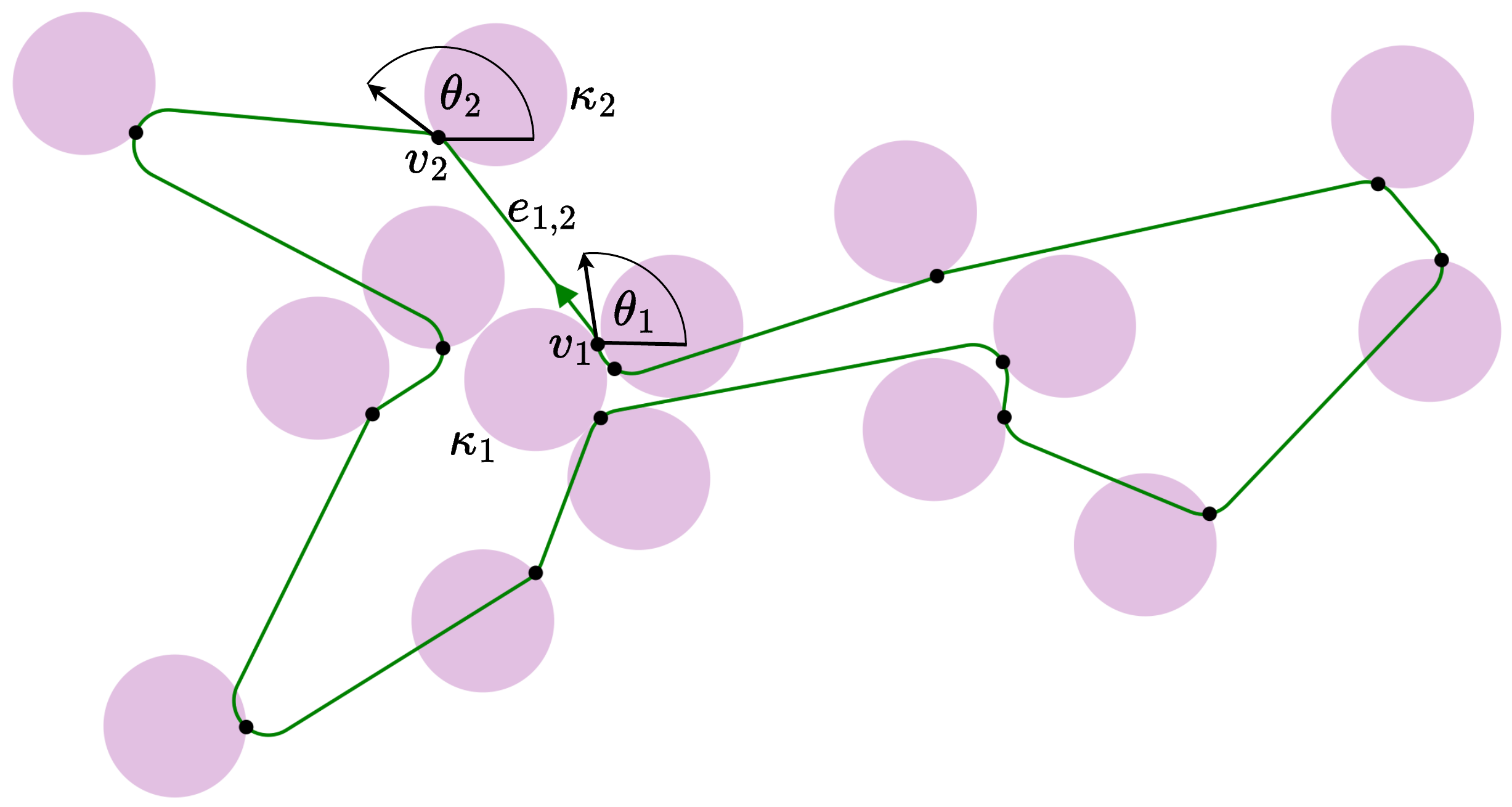}%
\label{fig:DTSPCP}}
\hfil
\subfloat[Optimal TSP solution, placement $\mathcal{K}$ given $r$]{\includegraphics[width=0.31\textwidth]{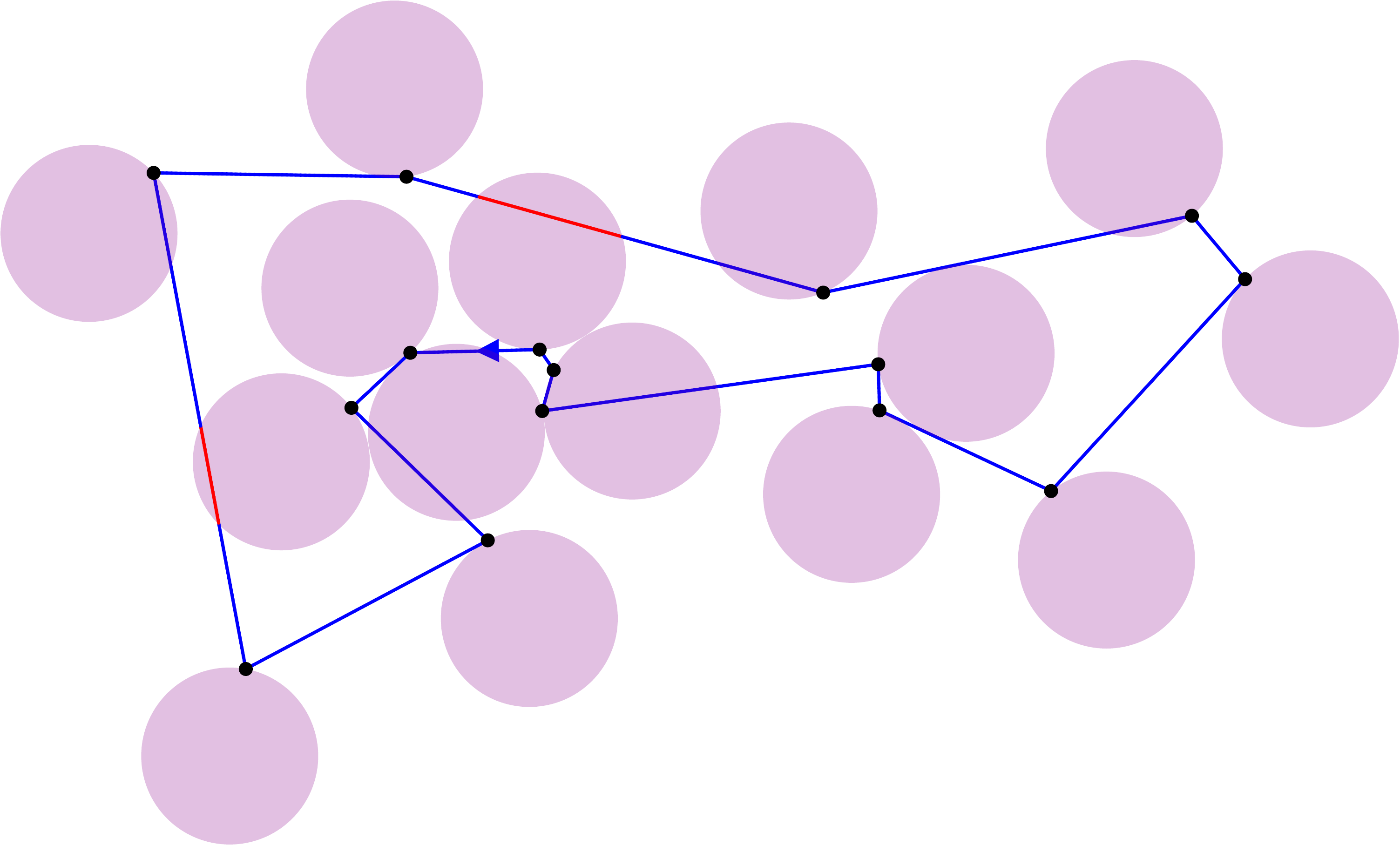}%
\label{fig:TSP}}
\caption{Example solutions ($v_i$ - $i^{th}$ node in cycle $c$, $e_{i,j}$ - edge from $v_i$ to $v_j$, $c_i$ - $i^{th}$ circle center, $\kappa_i$ - $i^{th}$ circle, $\theta_i$ - $i^{th}$ angular heading, $r$ - common radius). The paths start from $v_1$ and continue in the direction indicated by the arrow.}
\label{fig:example_solutions}
\end{figure*}

To describe the relevancy, take the open-pit mining application illustrated in Fig.~\ref{fig:drill_rig}, where the drill rig is tasked with drilling blast holes in predefined locations. The blast holes are then loaded with explosive materials in preparation for detonation.
The drilling generates a pile of excess material to be placed in the vicinity of each hole.
We illustrate the main criteria of the TSP-CP using this example: (1) creating a sufficiently large pile and maximizing clearance for the drill rig motion, i.e., placing the largest circles possible; (2) minimizing time and energy costs by finding the shortest tour; and (3) navigating among the obstacles created by the excess materials.
This task can be modeled as an instance of the TSP-CP whose one possible solution is depicted in Fig.~\ref{fig:TSPCP}.
A similar case can be envisaged in harvesting applications when a harvester needs to deposit the largest possible amount of crops in designated areas while respecting criteria 1--3 mentioned above. 

To address criteria 1--2 in the TSP-CP, we build upon the Weak Path Conforming Circle Placement Problem (WPCCP) introduced in~\cite{Kulich23}. In the WPCCP, the objective is to position the largest circles feasible on the nodes of a given tour. Unlike in the WPCCP, the tour is not given in the TSP-CP, thus enabling the placement of larger circles while simultaneously determining the optimal tour.

To account for criterion 3, i.e., navigating among circular obstacles in the TSP-CP, we employ the Travelling Salesperson Problem on self-deleting graphs (TSP-SD) formalism from~\cite{Carmesin23}. 
In a self-deleting graph, certain edges are deleted as a result of node visitation, allowing for representing the lack of reachability in a geometric extension of the graph with obstacles. 

In this paper, we propose a new problem-specific metaheuristic solver for the TSP-SD, which outperforms the generic solver used in~\cite{Carmesin23}.
Then, we introduce a novel heuristic TSP-CP solver, which combines the existing algorithm for the WPCCP~\cite{Kulich23} with our newly proposed TSP-SD solver.
Finally, we generalize the solver to accommodate the Dubins vehicle, thus introducing a new variant, the DTSP-CP\footnote{All source codes are available at \href{https://gitlab.ciirc.cvut.cz/imr/TSPSD/}{https://gitlab.ciirc.cvut.cz/imr/TSPSD/}}.

\section{Related Work}

The studied TSP-CP combines properties of several established optimization problems in various areas: Circle Packing Problems, Routing Problems, and Planning on temporal graphs. In this section, we provide an overview of these problems, describe their principal properties, and discuss their relevance to the TSP-CP.

Circle Packing Problems address the arrangement of circles in a container or a specified pattern~\cite{collins2003}.
One of the most relevant variants to the TSP-CP is the Uniform-sized Circle Packing Problem UCP2~\cite{castillo2008}, where the goal is to place a fixed number of uniform-sized, non-overlapping circles into a unit square and maximize their radius.
Other variants consider arbitrary-sized packings, different shapes of the encompassing container, minimization of its size, or even general geometries of the objects to be packed~\cite{Stoyan2017}.
However, Circle Packing Problems do not tackle the scenario where the placement is to be carried out by a vehicle. Thus, obstacle avoidance or trajectory optimization is not considered.

The trajectory planning aspect of TSP-CP is thoroughly studied in various Routing Problems, which focus on the mission and motion planning, while covering a predefined area or visiting given regions.
For example, the Traveling Salesperson Problem with Neighborhoods~\cite{Gentilini2013} tackles the problem of finding the shortest possible tour visiting a close neighborhood of fixed target locations, such as polygons~\cite{kulich2023}.
Another variant is the Close-Enough Traveling Salesperson Problem~\cite{Gulczynski2006}, where the goal is to get within a certain distance from the target location.
Thus, the task corresponds to visiting a fixed-radius circle, which is very close to the goal of the TSP-CP.
Both of these problems are also studied in environments with obstacles~\cite{kulich2023}~\cite{Deckerova2023} or with the Dubins vehicle model \cite{Vana2015}~\cite{Faigl2017}, which are common choices for many robotic applications.
However, unlike TSP-CP, none of these problems consider optimization of obstacle placement in time and space.

Probably the most challenging aspect of the TSP-CP is the evolution of the environment and the increasing number of obstacles over time.
Formulating the TSP on temporal graphs~\cite{Michail2016}, or time-evolving graphs~\cite{Sharma2021}, where the graph evolves in a predefined way in discrete steps, can achieve this behavior.
However, these formalisms ignore that, in the TSP-CP, obstacles arise as a consequence of the agent's actions rather than being a predefined function of time. Similarly, we cannot apply solutions from the Covering Canadian Traveller Problem~\cite{Liao2014}, where changes to the environment are not known in advance and unrelated to the agent's actions. 
A special case of the TSP-CP is studied in~\cite{Hellander2022}, where the vehicle is forbidden to physically pass through each node already visited.
Considering the dimensions of the vehicle, this corresponds to avoiding a circular obstacle centered at a given node.
Thus, the circle placement is not addressed.

The placement of circles along a given path was studied in~\cite{Kulich23} as the Weak Path Conforming Circle Placement Problem (WPCCP). In the WPCCP, circular obstacles are positioned along a TSP path to prevent collisions further along the route. However, the underlying TSP path is fixed. Additionally, the temporal aspect of the TSP-CP and tour optimization can be captured by the Traveling Salesperson Problem on self-deleting graphs (TSP-SD)~\cite{Carmesin23}, as the placed circular obstacles can be used to define the delete function in the TSP-SD. For these reasons, the TSP-CP problem includes the WPCCP and the TSP-SD as subproblems.

\section{Problem Statement}

In this section, we formally define the Travelling Salesperson Problem with Circle Placement, its Dubins variant, the concept of self-deleting graphs, and the Travelling Salesperson Problem on self-deleting graphs. 
Fig.~\ref{fig:example_solutions} shows solutions for instances of the TSP-CP and the DTSP-CP, using the problem formulation described in the current section.

\begin{prb} \label{def:TSP-CP}
Let $G = (V, E)$ be a simple, undirected, and complete Euclidean graph, where $V \subset \mathbb{R}^2$ and $|V| = n$.
The \emph{Travelling Salesperson Problem with Circle Placement (TSP-CP)} is to find a Hamiltonian cycle $c = \langle v_1, e_{1, 2}, v_2, .., v_n, e_{n, 1} \rangle$ and a set of circles $\mathcal{K} =\{\kappa_i\}_{i \in I}$, where $I = \{1,.., n\}$, $v_i \in V$, $e_{i, j} \in E$, $\kappa_i$ is a circle with center $c_i \in \mathbb{R}^2$ and radius $r_i \in \mathbb{R}$, such that the following requirements are met: \\
(C1) radii of all circles are equal - $\forall i \in I: r_i = r$, \\
(C2) $i^{th}$ point lies on $i^{th}$ circle - $\forall i \in I:  |v_i c_i| = r$, \\
(C3) intersection of any two circles is empty - $\forall i, j \in I, i \neq j: \kappa_i \cap \kappa_j = \emptyset$, \\
(C4) intersection of $i^{th}$ circle with the rest of the cycle is empty - $\forall i \in I: \kappa_i \cap \langle e_{i,i+1}, v_{i+1},..,v_n, e_{n, 1} \rangle = \emptyset$, \\
(O1) radius of circles $r$ is maximal, \\
(O2) given $r$, cycle cost $\sum_{i = 1}^{n-1}|e_{i, i + 1}| + |e_{n, 1}|$ is minimal.
\end{prb}

The definition of the TSP-CP problem directly generalizes the Weak Path-Conforming Circles Placement Problem (WPCCP) from~\cite{Kulich23}, where an input path is given and not subject to optimization.
Note that (O1) criterion maximizing $r$ is the primary objective, analogously to the WPCCP. 
The secondary objective (O2) then prioritizes the shorter among two solutions, given that the $r$ is equal. 
In addition, the TSP-CP can be easily reformulated for the Dubins vehicle model.

\begin{prb} \label{def:DTSP-CP}
Let $V$ $\subset \mathbb{R}^2$ be a set of points, $|V| = n$, $r_{dub} \in \mathbb{R}$. 
The \emph{Dubins Travelling Salesperson Problem with Circle Placement (DTSP-CP)} is to find a a Hamiltonian cycle $c = \langle v_1, \theta_1, e_{1, 2}, v_2, \theta_2, .., v_n, \theta_n, e_{n, 1} \rangle$ and a set of circles $\mathcal{K} =\{\kappa_i\}_{i \in I}$, where $I = \{1,.., n\}$, $v_i \in V$, $\theta_i \in [0, 2\pi)$ is angular heading from $v_i$, $e_{i,j}$ is the shortest Dubins maneuver from $(v_i, \theta_i)$ to $(v_j, \theta_j)$ with turning radius $r_{dub}$, and $\kappa_i$ is a circle with center $c_i \in \mathbb{R}^2$ and radius $r_i \in \mathbb{R}$.
The constraints and optimality criteria are identical with Problem~\ref{def:TSP-CP}.
\end{prb}

Finally, let us define self-deleting graphs and the Traveling Salesperson Problem on self-deleting graphs (TSP-SD). The definitions are adopted from~\cite{Carmesin23}.

\begin{dfn}
A \emph{self-deleting graph $S$} is a tuple $S = (G, f)$ where $G = (V, E)$ is a simple, directed graph and $f:V \xrightarrow{} 2^E$ is a \emph{delete function}.
If a vertex $v$ is \emph{processed}, edges $f(v)$ are deleted from $G$.
Given a set of vertices $X \subset V$, the \emph{residual graph} $G_{X}$ of $S$ after processing $X$ is defined as $G_{X} = G \backslash \bigcup_{v \in X} f(v)$.
Then, a path $P = \langle p_1, p_2, .., p_x \rangle$ in $S$ is \emph{f-conforming}, if for every $1 \leq i < x $ the edge $e_{i, i+1}$ is in the residual graph $G_{\{p_1, p_2, .., p_i\}}$.
Similarly, a sequence of vertices $c = \langle p_1, .., p_n \rangle$ is an $f$-conforming \emph{Hamiltonian cycle}, if $\langle p_1, .., p_n \rangle$ is an $f$-conforming path and the edge $e_{n, 1} \in G_{V}$.
Finally, let us define the $cost(c) = \sum_{i=1}^{n-1}|e_{i, i+1}| + |e_{n,1}|$, where $|e_{i, j}|$ is the length of $e_{i, j}$.
\end{dfn}

\begin{prb}\label{def:TSP-SD}
Given $S = (G, f)$, the \emph{Travelling Salesperson Problem on self-deleting graphs (TSP-SD)} is to find the shortest f-conforming Hamiltonian cycle on $S$.
\end{prb}

\section{Methods}

In this section, we describe in detail all proposed solvers starting from the subproblems involved. First, we present a novel metaheuristic solver for the TSP-SD. Next, we detail a heuristic solver designed for the TSP-CP with a fixed radius and a heuristic solver for the TSP-CP. 
Finally, we describe the adaptations of both TSP-CP solvers for the DTSP-CP.
All proposed solvers are heuristic and do not guarantee optimality.
However, the TSP-SD solver and TSP-CP solver guarantee finding a feasible solution.

\subsection{GRASP solver for the TSP-SD}

The greatest challenge of TSP-CP is to satisfy constraint C4.
As the cycle $c$ is traversed, the circles $\mathcal{K}$ are placed sequentially, one with each visited node, and represent new obstacles.
Thus, the set of available edges in $E$ is reduced with each step.
This dynamic property of the TSP-CP can be captured by the TSP-SD.
This section presents the proposed solver for the TSP-SD, which implements the Greedy Randomized Adaptive Search Procedure~\cite{Resende2016}.

Solving TSP-CP requires repeatedly solving TSP-SD, and the generic solver~\cite{woller2022} used in~\cite{Carmesin23} fails to converge to good-quality solutions for large instances (see results in Tab.~\ref{tab:GRASP}).
Thus, we propose a more efficient problem-specific solver, that leverages instance properties unique to the TSP-SD. 
Within the GRASP metaheuristic, it combines the backward search procedure from~\cite{Carmesin23} for the initial solution generation and several new custom-tailored local search operators.

\subsubsection{GRASP metaheuristic}

\begin{algorithm}[ht]
\caption{GRASP metaheuristic} \label{alg:GRASP}
\begin{algorithmic}[1]
\REQUIRE Self-del. graph $S = (G, f)$, TSP tour $c_{TSP}$ on $G$
\STATE $c_{TSPSD}^* = \varnothing$
\WHILE{$\mathtt{!stop()}$}
    \STATE $c_{TSPSD} = \mathtt{backward\_search(}S, c_{TSP}\mathtt{)}$ \\            \label{alg:GRASP_init}
    \IF {$\mathtt{is\_valid(}c_{TSPSD}\mathtt{)}$}                          
        \STATE $\mathtt{local\_search(}S, c_{TSPSD}\mathtt{)}$                      \label{alg:GRASP_ls}
        \IF {$\mathtt{cost(}c_{TSPSD}\mathtt{)} < \mathtt{cost(}c_{TSPSD}^{*}\mathtt{)}$}
            \STATE $c_{TSPSD}^{*} = c_{TSPSD}$
        \ENDIF                                                              \label{alg:GRASP_eval}
    \ENDIF
\ENDWHILE
\STATE \textbf{return} Hamiltonian cycle $c_{TSPSD}^{*}$ on $S$ \label{alg:GRASP_ret}
\end{algorithmic}
\end{algorithm}

Alg.~\ref{alg:GRASP} describes the GRASP metaheuristic. 
The algorithm takes as input a self-deleting graph $S = (G, f)$ and optionally a TSP cycle $c_{TSP}$, i.e., a TSP solution on $G$.
The delete function $f$ is created based on a circle placement around $c_{TSP}$ in the proposed TSP-CP algorithm, so the $c_{TSP}$ is likely to have only a few collisions (see Fig.~\ref{fig:TSP}).
Then, the following procedure is repeated until a custom stop condition is met.
An initial solution $c_{TSPSD}$ is created (line~\ref{alg:GRASP_init}), using the backward search procedure proposed in~\cite{Carmesin23}.
In the backward search, $c_{TSP}$ determines the expansion order while building $c_{TSPSD}$.
It is beneficial for $c_{TSPSD}$ to be as similar to the input $c_{TSP}$ as possible, as the $c_{TSP}$ can be of good quality or even optimal on $G$ and mostly without collisions on $S$.
The resulting $c_{TSPSD}$, if found, is guaranteed to be $f$-conforming and starts at the same node as $c_{TSP}$.
If $c_{TSP}$ is not used, the backward search expands semi-greedily from a random start point.
Then, the local search attempts to improve the cost of $c_{TSPSD}$, while keeping it $f$-conforming (lines~\ref{alg:GRASP_ls}-\ref{alg:GRASP_eval}).
Finally, the best solution $c_{TSPSD}^{*}$ found is returned (line~\ref{alg:GRASP_ret}).

\subsubsection{Local search}

\begin{figure}[t]
    \centering
    \subfloat{\includegraphics[width=0.3\textwidth]{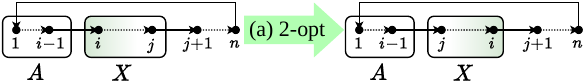}%
    \label{fig:twoOpt}} \\
    \subfloat{\includegraphics[width=0.4\textwidth]{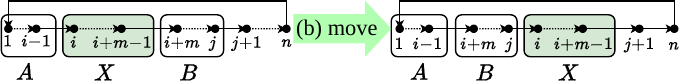}%
    \label{fig:move}} \\
    \subfloat{\includegraphics[width=0.48\textwidth]{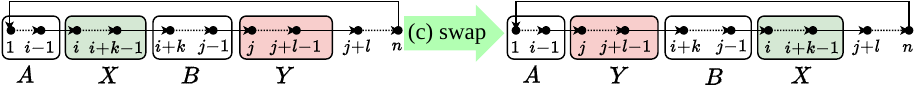}%
    \label{fig:swap}}
    \caption{Local search operators for the TSP-SD.}
    \label{fig:ls_operators}
\end{figure}

The goal of the local search is to reach a local optimum w.r.t. multiple neighborhoods - sets of solutions reachable by applying a local search operator to a current solution.
We propose three operators: 2-opt, move, and swap, where each of them defines a different neighborhood.
The operators assume the asymmetric TSP-SD ($e_{i,j} \neq e_{j,i}$) and do not change the first and last node of the modified solution.
Thus, the start and end can be fixed. 
On a higher level, the operators are controlled by the Cyclic Variable Neighborhood Descent (CVND) heuristic~\cite{Duarte2018}, which exhaustively searches all neighborhoods in fixed order until no improvement is obtained in any of them.
The rest of this section presents individual operators. 
Fig.~\ref{fig:ls_operators} shows the solution before and after applying each operator for a given pair of indices $i, j$.
The moved or inverted solution parts are highlighted in color.

The operators are designed so that both the validity of the solution and the cost are evaluated incrementally.
This significantly enhances the scalability of the proposed solver.
If an operator is applicable for all pairs of indices $i,j \in \{1,..,n\}$, then a naively implemented exhaustive search of the corresponding neighborhood has a complexity of $\mathcal{O}(n^3)$, $\mathcal{O}(n^3 \overline{f(p)})$ in case of cost and validity evaluation, respectively. 
Here, $\overline{f(p)}=\sum_{i=1}^{n} f(p_i)/n$ is the average number of deleted edges per node. 
When evaluated incrementally, the complexities can be reduced to $\mathcal{O}(n^2)$ and $\mathcal{O}(n^2 \overline{f(p)})$.

\subsubsection{2-opt}

The 2-opt operator (Fig.~\ref{fig:twoOpt}) is described in Alg.~\ref{alg:2-opt}.
Given indices $i, j$, 2-opt inverts a subsequence $x=\langle p_i,.., p_j\rangle$ of a cycle $c = \langle p_1,.., p_n \rangle$.
The operator iterates over all pairs of indices $i, j$ and applies the first improving pair found.
Let us define the set of nodes in $x$ as $X=\{p_i, .., p_j\}$, the set of nodes before $x$ as $A=\{p_1,..,p_{i-1}\}$, and the set of newly added edges in inverted $x$ as $E_X = \{e_{j,j-1},.., e_{i+1, i}\}$.
Then, let $E_{XA}$ contain edges that are in $E_X$, but are not in the residual graph $G_A$ - newly added edges deleted by some of the nodes in $A$.
Finally, we can formulate several rules for the 2-opt-modified cycle $c_{cand} = \langle p_1,..,p_j,..,p_i,..p_n \rangle$ to be $f$-conforming: \\
(i) $X$ must not delete edges in $E_X$ - incrementally checked for edge $e_{i+1,i}$ at line~\ref{alg:2-opt-rule1} of Alg.~\ref{alg:2-opt}. \\
(ii) $A$ must not delete edges in $E_X$ - currently added $e_{i+1,i}$ is checked at line~\ref{alg:2-opt-rule2}, previously deleted edges are stored in $E_{XA}$ (line~\ref{alg:2-opt-EXA_add}) and checked at line~\ref{alg:2-opt-rule345}. 
As $A$ becomes smaller (line~\ref{alg:2-opt-A_update}), $E_{XA}$ must be updated (line~\ref{alg:2-opt-EXA_update}). \\
(iii) $e_{i-1, j}$ must not be deleted by $A$ (line~\ref{alg:2-opt-rule345}). \\
(iv) $e_{i, j+1}$ must not be deleted by $X$ and $A$ (line~\ref{alg:2-opt-rule345}).

If all rules are satisfied, $cost(c_{cand})$ can be evaluated in constant time (line~\ref{alg:2-opt_cost-2}).
If the move is improving, the current $c_{cand}$ is returned (line~\ref{alg:2-opt_return_1}).
Otherwise, the original $c$ is returned after testing all pairs of $i,j$ (line~\ref{alg:2-opt_return_2}).

\begin{algorithm}[ht]
\caption{2-opt} \label{alg:two-opt}
\begin{algorithmic}[1]
\REQUIRE $S=(G, f)$, Ham. cycle $c = \langle p_1, .., p_n \rangle$ on $S$
\FOR{$j = n-1$ to $3$}
    \STATE $X=\{p_{j}\}, E_{XA} = \{\}$
    \FOR{$i = j - 1$ to $2$}
        \STATE $A=\{p_1,..,p_{i-1}\}$   \label{alg:2-opt-A_update}
        \IF {$e_{i+1, i} \in G_X$} \label{alg:2-opt-rule1}                                   
            \STATE $X = X \cup \{p_{i}\}$, $E_{XA} = E_{XA} \setminus G_A$ \\ \label{alg:2-opt-AX_update} \label{alg:2-opt-EXA_update}
            \IF {$e_{i+1, i} \in G_A$} \label{alg:2-opt-rule2} 
                \IF {$e_{i-1, j} \in G_{A} \land e_{i, j+1} \in G_{A \cup X} \land |E_{XA}| = 0$} \label{alg:2-opt-rule345} 
                    \IF {$\mathtt{eval_{2-opt}(}c, i, j\mathtt{)} < \mathtt{cost(}c\mathtt{)}$} \label{alg:2-opt_cost-2}
                        \STATE \textbf{return} $\mathtt{apply\_two\_opt(}c, i, j\mathtt{)}$ \label{alg:2-opt_return_1}
                    \ENDIF
                \ENDIF
            \ELSE
                \STATE $E_{XA} = E_{XA} \cup e_{i+1, i}$ \label{alg:2-opt-EXA_add}
            \ENDIF
        \ELSE
            \STATE \textbf{break}
        \ENDIF
    \ENDFOR
\ENDFOR
\STATE \textbf{return} unchanged cycle $c$ \label{alg:2-opt_return_2}
\end{algorithmic} \label{alg:2-opt}
\end{algorithm}


\subsubsection{Move}

The move operator (Fig.~\ref{fig:move}) attempts to move a subsequence $x=\langle p_i,..,p_{i+m-1} \rangle$ of fixed length $m$, starting at index $i$, to a better position $j$ in cycle $c=\langle p_1,.., p_n \rangle$.
Alg.~\ref{alg:move} describes one branch of the operator - move-right, where $x$ is moved only to the right from $i$.
The second branch functions analogously.
Let $X$ the set of nodes in $x$, $E_X$ the edges in $x$, and $A$ the set of nodes in $c$ before $x$ (lines~\ref{alg:move_init_1}-\ref{alg:move_init_2}).
Then, $B$ is the set of nodes between $x$ and $j$ (line~\ref{alg:move_init_3}).
For each tested combination of $i,j$, the following three conditions must be satisfied for the new edges in $c_{cand} = \langle p_1,..,p_{i-1}, p_{i+m}, .. p_{j}, p_i, .., p_{i+m-1}, p_{j+1}, .., p_n\rangle$: \\
(i) $e_{i-1, i+m}$ must not be deleted by $A$ (line~\ref{alg:move_rule1}), \\
(ii) $e_{j,i}$ must not be deleted by $A, B$ (line~\ref{alg:move_rule23}), \\
(iii) $e_{i+m-1, j+1}$ must not be deleted by $A, B, X$ (line~\ref{alg:move_rule23}), \\
plus one for those edges that were moved right: \\
(iv) $E_X$ must not be deleted by $B$ (line~\ref{alg:move_rule4}). \\
When (i)-(iv) are satisfied, the cost of $c_{cand}$ is evaluated in constant time.
If an improving move is found, it is applied and returned.
Otherwise, $c$ remains unchanged (line~\ref{alg:move_eval}-\ref{alg:move_return}).


\begin{algorithm}[t]
\caption{move-right}
\begin{algorithmic}[1]
\REQUIRE $S=(G, f)$, $c = \langle p_1, .., p_n \rangle$, moved seq. length $m$ \\
\FOR{$i = 2$ to $n-m-1$}
    \STATE $A = \{p_1,..,p_{i-1}\}, X = \{p_i,..,p_{i+m-1} \}$    \label{alg:move_init_1}
    \STATE $E_X = \{e_{i, i+1},..,e_{i+m-2, i+m-1}\}$              \label{alg:move_init_2}
    \IF{$e_{i-1, i+m} \in G_A$}         \label{alg:move_rule1}    
        \FOR{$j = i + m$ to $n-1$}
            \STATE $B = \{p_{i+m},..,p_j \}$   \label{alg:move_init_3}
            \IF {$E_X \not\subset G_B$} \label{alg:move_rule4}
                \STATE \textbf{break} 
            \ENDIF
            \IF{$e_{j, i} \in G_{A \cup B} \land e_{i+m-1, j+1} \in G_{A \cup B \cup X}$} \label{alg:move_rule23}
                \IF{$\mathtt{eval_{mr}(}c, i, j, m\mathtt{)} < \mathtt{cost(}c\mathtt{)}$} \label{alg:move_eval}
                    \STATE \textbf{return} $\mathtt{apply\_move\_right(}c,i,j,m\mathtt{)}$  \label{alg:move_return_1}
                \ENDIF
            \ENDIF
        \ENDFOR
        \ENDIF
\ENDFOR
\STATE \textbf{return} unchanged cycle $c$ \label{alg:move_return}
\end{algorithmic} \label{alg:move}
\end{algorithm}

\subsubsection{Swap}

The swap operator (Fig.~\ref{fig:swap}), described in Alg.~\ref{alg:swap}, attempts to swap two subsequences $x = \langle p_i,..,p_{i+k-1} \rangle, y = \langle p_j,..,p_{j+l-1}\rangle$ of fixed lengths $k, l$.
Assuming $i < j$, the operator is not symmetric, thus $\mathtt{swap}$-$\mathtt{asym}(S,c,k,l)$ and $\mathtt{swap}$-$\mathtt{asym}(S,c,l,k)$ must both be run if $k \neq l$.
Note that the swap operator cannot be replaced by running the move operator sequentially with parameters $k,l$, as the intermediate solution may be infeasible due to the delete function $f$.
Let us denote the set of nodes in $c$ before $x$ as $A=\{p_1,..,p_{i-1}\}$, nodes in $x$ as $X$, nodes between $x$ and $y$ as $B = \{p_{i+k},..,p_{j-1}\}$ and nodes in $y$ as $Y$ (line~\ref{alg:swap_init_1}).
If $i+k=j, B = \emptyset$.
For each tested combination of $i,j$, the following six conditions must be satisfied so that $c_{cand} = \langle p_1,.., p_j,..,p_{j+l-1},.., p_i, ..,p_{i+k-1},..,p_n \rangle$ is $f$-conforming: \\
(i) nodes in $Y$ are shifted to the left, so they must not delete edges $E_B$ in $B$ (line~\ref{alg:swap_cond2}). \\
All newly added edges must not be deleted (line~\ref{alg:move_cond1}): \\
(ii) $e_1 = e_{i-1, j}$ by $A$, 
(iii) $e_2$ by $A, Y$, 
(iv) $e_3$ by $A, Y, B$, \\
(v) $e_4 = e_{i+k-1, j+l}$ by $A, Y, B, X$. \\
(vi) edges $E_X$ in $x$ are shifted to the right, so they must not be deleted by $B$ (line~\ref{alg:move_cond1}). 
If $B = \emptyset$, $e_2 = e_3$.
If all conditions are met, $cost(c_{cand})$ can be evaluated in constant time.
If an improving move is found, it is applied and returned (line~\ref{alg:swap_return1}).

\begin{algorithm}[t]
\caption{swap-asym}
\begin{algorithmic}[1]
\REQUIRE $S=(G, f)$, $c = \langle p_1, .., p_n \rangle$, swapped seq. lengths $k, l$
\FOR{$j = n-l$ to $k + 2$}
    \STATE $Y = \{p_j,..,p_{j+l-1}\}$
    \FOR{$i = j-k$ to $2$}
        \STATE $A = \{p_1,..,p_{i-1}\}$, $X = \{p_i,..,p_{i+k-1}\}$, $B = \{p_{i+k},.., p_{j-1}\}$, $E_X = \{e_{i, i+1},..,e_{i+k-2, i+k-1}\}$ \label{alg:swap_init_1}
        \STATE $e_1 = e_{i-1, j}, e_4 = e_{i+k-1, j+l}$
        \IF{$E_B \not\subset G_Y$} \label{alg:swap_cond2}
            \STATE $\textbf{break}$
        \ENDIF
        \IF{$i+k=j$}
            \STATE $e_2 = e_3 = e_{j+l-1, i}$
        \ELSE
            \STATE $e_2 = e_{j+l-1, i+k}, e_3 = e_{j-1, i}$            
        \ENDIF
        \IF{$e_1 \in G_A \land e_2 \in G_{A \cup Y} \land e_3 \in G_{A \cup Y \cup B} \land e_4 \in G_{A \cup Y \cup B \cup X} \land E_X \subset G_B \land E_X \subset G_Y$} \label{alg:move_cond1}
            \IF{$\mathtt{eval_{sa}}(c, i, j, k, l)$}
                \STATE \textbf{return} $\mathtt{apply\_swap\_asym}(c, i, j, k, l)$ \label{alg:swap_return1}
            \ENDIF
        \ENDIF        
    \ENDFOR
\ENDFOR
\STATE \textbf{return} unchanged $c$ \label{alg:swap_return2}
\end{algorithmic} \label{alg:swap}
\end{algorithm}

\subsection{TSP-CP solver - fixed radius}

The proposed solver for the TSP-CP with a fixed radius is described in Alg.~\ref{alg:TSP-CP_fixed_r}.
The input is a Euclidean graph $G=(V, E)$, a radius $r$ and a TSP cycle $c_{TSP}$.
If the instance is feasible, the output is a valid TSP-CP Hamiltonian cycle and a set of circles $\mathcal{K}$ of the given radius $r$.
First, a placement of circles $\mathcal{K}$ along $c_{TSP}$ is found using the Local search algorithm for a relaxed problem: the WPCCP with fixed radius~\cite{Kulich23} (line~\ref{alg:TSPCPf_WPCCP}). 
For each node $v_i \in c_{TSP}$, $c_{TSP}=\langle v_1,e_{1,2}..,v_n,e_{n,1}\rangle$, this algorithm samples a set of candidate circles $\kappa_i^c$, discards those candidate circles that are in collision with the rest of the cycle $\langle v_i,e_{i,i+1}..,v_n,e_{n,1}\rangle$ and then finds a least-colliding placement $\mathcal{K}$ using a simple local optimization procedure.
Here, the algorithm is configured to treat collisions between the edges of the cycle and the circles as a penalized soft constraint.
Collisions between the circles themselves are not allowed.
If the algorithm returns a circle placement $\mathcal{K}$, it is guaranteed to respect constraints (C1), (C2) and (C3) of Problem~\ref{def:TSP-CP}.
The constraint (C4) may not be satisfied.
If it is, $c_{TSP}$ and $\mathcal{K}$ represent a valid solution to the TSP-CP with a fixed radius and are returned (line~\ref{alg:TSPCPf_return1}). 
The $cost(c_{TSP})$ is optimal for TSP-CP if it is optimal for TSP.
Both problems are solved on the same graph $G$, but some edges may be removed in the TSP-CP.
Thus, the optimal TSP-CP solution cannot be better than the TSP one.

Otherwise, $\mathcal{K}$ satisfies (C1), (C2), and (C3), but the initial $c_{TSP}$ violates (C4), as illustrated in Fig.~\ref{fig:TSP}.
Then, a TSP-SD instance is generated for $G=(V, E)$ and $\mathcal{K}$ (line~\ref{alg:TSPCPf_TSPSDgen}) in the following fashion: $\forall e_i \in E, v_j \in V: f(v_j) \gets f(v_j) \cup e_i \iff k_j \cap e_i \neq \emptyset$.
In other words, an edge $e_i$ is added to the delete function $f(v_j)$, if $e_i$ intersects the circle $k_j \in \mathcal{K}$ corresponding to the node $v_j$.
The resulting TSP-SD instance $S=(G, f)$ is then solved using the proposed TSP-SD GRASP solver (line~\ref{alg:TSPCPf_GRASP}).
In this phase, the circle placement $\mathcal{K}$ is fixed, and the TSP-SD solver attempts to rearrange the cycle $c_{TSP}$ to satisfy (C4) and minimize (O2) to a local optimum.
If it succeeds, the resulting $c_{TSPSD}$ is a valid TSP-CP cycle.
The TSP-SD solver uses an exact construction procedure, which is run 20 times with a time limit of 20 seconds.
If it fails to find a valid initial solution, the TSP-SD subproblem is considered infeasible.

\begin{algorithm}[t]
\caption{TSP-CP solver - fixed radius}
\label{alg:TSP-CP_fixed_r}
\begin{algorithmic}[1]
    \REQUIRE Euclidean graph $G$, radius $r$, TSP cycle $c_{TSP}$
    \STATE \blue{$\mathcal{K} = \mathtt{place\_circles\_soft}(c_{TSP}, r)$}    \label{alg:TSPCPf_WPCCP}
    \IF{$\mathtt{is\_valid\_TSPCP}(c_{TSP}, \mathcal{K})$}
        \STATE $\textbf{return } c_{TSP}, \mathcal{K}$  \label{alg:TSPCPf_return1}
    \ELSIF{$\mathcal{K} \neq \emptyset$}
        \STATE \blue{$f = \mathtt{generate\_TSPSD}(G, \mathcal{K})$}   \label{alg:TSPCPf_TSPSDgen}
        \STATE $c_{TSPSD} = \mathtt{GRASP}(G, f, c_{TSP})$  \label{alg:TSPCPf_GRASP}
        \IF{$c_{TSPSD} \neq \varnothing$}
            \STATE \textbf{return} TSP-CP cycle $c_{TSPSD}$, circles $\mathcal{K}$
        \ELSE
            \STATE \textbf{return} $\varnothing$
        \ENDIF
    \ENDIF
\end{algorithmic}
\end{algorithm}

\subsection{TSP-CP solver}

The proposed approach to solve the TSP-CP is presented in Alg.~\ref{alg:TSPCP}. 
The input to the algorithm is a Euclidean graph $G=(V,E)$, and the output is a placement of circles $\mathcal{K}$ of maximum radius $r$ and an $f$-conforming TSP-CP Hamiltonian cycle $c$ of minimal length.
First, the optimal TSP solution $c_{TSP}$ for $G$ is found using the exact Concorde TSP solver~\cite{Cook20} (line~\ref{alg:TSPCP_TSP}).
Then the Interval Bisection Algorithm for the Weak Path Conforming Circles Placement Problem (WPCCP) from~\cite{Kulich23} is run, given $c_{TSP}$ (line~\ref{alg:TSPCP_WPCCP}).
This algorithm guarantees finding a placement $\mathcal{K}$, that satisfies all constraints (C1) to (C4) and maximizes the radius $r$ (criterion (O1)), given a fixed cycle $c_{TSP}$.
It also returns the upper bound estimate $ub$ for $r$.
However, the radius $r$ can be further increased if we relax the WPCCP to the TSP-CP and allow the input cycle $c_{TSP}$ to change.
For this purpose, the bisection algorithm is adapted for the TSP-CP in the rest of Alg.~\ref{alg:TSPCP} (lines~\ref{alg:TSPCP_bisection_start}-\ref{alg:TSPCP_bisection_end}).
The initial interval is given by $ub$ and $r$ from the previous step. 
In each loop of the interval bisection, a new target radius $r_{cand}>r$ is calculated (line~\ref{alg:TSPCP_r_cand}).
Then, given $G, r_{cand}$ and $c_{TSP}$, the TSP-CP solver for fixed radius is run in an attempt to find a new $f$-conforming TSP-CP cycle $c_{cand}$ and new circle placement $\mathcal{K}_{cand}$ (line~\ref{alg:TSPCP_fixed}).
According to the result, the bounds $lb, ub$ are adjusted (lines~\ref{alg:TSPCP_adjust_1}-\ref{alg:TSPCP_adjust_2}).
The algorithm teminates once the bounds difference $ub-lb$ is within a predefined limit $\epsilon=0.1$ (line~\ref{alg:TSPCP_bisection_start}) and returns the best values of $c, \mathcal{K}$ and $r$ (line~\ref{alg:TSPCP_return}).

\begin{algorithm}[t]
\caption{TSP-CP solver}
\label{alg:TSPCP}
\begin{algorithmic}[1]
    \REQUIRE Euclidean graph $G$
    \STATE \blue{$c_{TSP} = \mathtt{solve\_TSP}(G)$}                                   \label{alg:TSPCP_TSP}
    \STATE \blue{$\mathcal{K}, r, ub = \mathtt{interval\_bisection}(c_{TSP})$}         \label{alg:TSPCP_WPCCP}
    \STATE $lb = r, c = c_{TSP}$
    \WHILE{$ub - lb > \epsilon$}                                                \label{alg:TSPCP_bisection_start}
        \STATE $r_{cand} = \frac{lb + ub}{2}$                                   \label{alg:TSPCP_r_cand}
        \STATE $c_{cand}, \mathcal{K}_{cand} = \mathtt{TSPCP\_fixed\_r}(G, r_{cand}, c_{TSP})$  \label{alg:TSPCP_fixed}
        \IF{$c_{cand} \neq \varnothing$}                                            \label{alg:TSPCP_adjust_1}
            \STATE $lb = r_{cand}$, $c = c_{cand}$, $K = K_{cand}$, $r=r_{cand}$
        \ELSE
            \STATE $ub = r_{cand}$
        \ENDIF                                                                      \label{alg:TSPCP_adjust_2}
    \ENDWHILE                                                                   \label{alg:TSPCP_bisection_end}
    \STATE \textbf{return} TSP-CP cycle $c$, circles $\mathcal{K}$, radius $r$ \label{alg:TSPCP_return}
\end{algorithmic}
\end{algorithm}

\subsection{Dubins TSP-CP solver}

The proposed approach for the DTSP-CP directly adapts the TSP-CP solvers described in Alg.~\ref{alg:TSP-CP_fixed_r} and~\ref{alg:TSPCP} to the Dubins vehicle model~\cite{SHKEL2001}. The Dubins variant is principally identical to the Euclid variant.
However, several components need to be replaced or adjusted.
All necessary changes are discussed in this section and marked in blue in Alg.~\ref{alg:TSP-CP_fixed_r} and~\ref{alg:TSPCP}.
A Dubins maneuver is given by a pair of configurations $(x, y, \theta)$ and a turning radius $r_{dub}$, where $x, y \in \mathbb{R}^2$ are the coordinates of the maneuver endpoints, $\theta \in [0, 2\pi)$ are angular headings at each endpoint.
We only consider the shortest maneuver, which is unique~\cite{SHKEL2001}.

The first modification is needed on line~\ref{alg:TSPCPf_TSPSDgen} of Alg.~\ref{alg:TSP-CP_fixed_r}, where a TSP-SD instance is generated, given graph $G$ and circle placement $\mathcal{K}$.
Instead of using the Euclidean graph $G=(V,E)$ as input, a graph $G_{dub}=(V_{dub}, E_{dub})$ is created, based on the cycle $c_{DTSP} = \langle v_1, \theta_1, e_{1,2},..,v_n, \theta_n, e_{n,1} \rangle$, which is a DTSP solution on $G$.
The nodes are defined as $V_{dub} = \{v_i \in c_{DTSP}\}$, where $v_i = (x_i, y_i, \theta_i)$.
In other words, the headings $\theta$ are part of the node coordinates and taken from $c_{DTSP}$. 
Therefore, the headings are fixed in $V_{dub}$.
The edges $E_{dub}$ are then simply the shortest Dubins maneuvers between all pairs of nodes in $V_{dub}$, and the TSP-SD delete function $f$ is given by collisions between circles $\kappa_i \in \mathcal{K}$ and Dubins maneuvers in $E_{dub}$.

The second modification is required on line~\ref{alg:TSPCP_TSP} of Alg.~\ref{alg:TSPCP}, where we generate the initial DTSP cycle $c_{DTSP}$.
It corresponds to the need to select an appropriate heading $\theta_i$ in each node.
This is achieved by transforming the problem into the Generalized Traveling Salesperson Problem (GTSP), where the goal is to plan the shortest cycle possible, visiting exactly one element from each set $V_i$, $i \in \{1,..,n\}$.
Here, $V_i = \{(v_i, \theta_1),..,(v_i, \theta_k)\}$, where the headings $\theta_j$, $j \in \{1,..,k\}$ are uniformly sampled for each $v \in V$.
This subproblem is solved using the Generalized Large Neighborhood Search (GLNS)~\cite{Smith2016GLNS} metaheuristic solver for the GTSP.

The third modification is required on line~\ref{alg:TSPCPf_WPCCP} of Alg.~\ref{alg:TSP-CP_fixed_r}, where the Local search algorithm for the WPCCP~\cite{Kulich23} is used to find a circle placement $\mathcal{K}$, given cycle $c_{TSP}$.
For each node $v_i \in c_{TSP}$, $c_{TSP}=\langle v_1,e_{1,2}..,v_n,e_{n,1}\rangle$, this algorithm samples a set of candidate circles $\kappa_i^c$, penalizes or discards those candidate circles that are in collision with the rest of the cycle $\langle v_i,e_{i,i+1}..,v_n,e_{n,1}\rangle$ and then finds a least-colliding placement $\mathcal{K}$.
The only difference in the Dubins variant is that collisions must be checked between Dubins maneuvers in $c_{DTSP}$ and circles $\kappa_i^c$, which can be easily added to the implementation. 
The same Local search algorithm is then used within the Interval Bisection Algorithm for the WCCP~\cite{Kulich23}, used in line~\ref{alg:TSPCP_WPCCP} of Alg.~\ref{alg:TSPCP}, where the colliding candidate circles $\kappa_i^c$ are discarded.

\section{Experiments}

All proposed algorithms are experimentally evaluated.
First, the GRASP solver for the TSP-SD is benchmarked against the generic solver employed in~\cite{Carmesin23}.
Second, the TSP-CP solver is compared with the WPCCP solver from~\cite{Kulich23}.
Finally, the behavior of the DTSP-CP solver variant is studied on a small dataset for several different Dubins turning radii and compared with the TSP-CP solutions.
All experiments were carried out on a dedicated machine with Ubuntu 18.04, Intel Core i7-7700, including those from~\cite{Kulich23, Carmesin23}.
The gap values presented are calculated as $gap = 100(\frac{score}{ref} - 1)$, where $score$ is some performance metric of our algorithm (e.g. mean solution cost), and $ref$ is a reference method performance metric (e.g. best-known solution so far).
Similarly, the standard deviation $\sigma$ of $score$ is also expressed relative to $ref$ as $gap(\sigma)=100\frac{\sigma}{ref}$.

\begin{table*}[]
\caption{TSP-CP results - datasets from~\cite{Kulich23}: hexagons, meshes; new datasets: noisy hexagons, squares}
\label{tab:TSP-CP_1}
\centering
\begin{tabular}{p{0.8cm}rr|r|rrr|r||p{0.85cm}rr|r|rrr|r}
\hline
\multicolumn{3}{r|}{reference solver~\cite{Kulich23}} & \multicolumn{5}{c||}{proposed TSP-CP solver} & \multicolumn{3}{r|}{reference solver~\cite{Kulich23}} & \multicolumn{5}{c}{proposed TSP-CP solver} \\
\multicolumn{2}{r}{BKS} & $\overline{\mbox{time}}$ & \multicolumn{1}{r}{r\textsubscript{max}} & \multicolumn{3}{c}{gap values (\%)} & $\overline{\mbox{time}}$ & \multicolumn{2}{r}{BKS} & $\overline{\mbox{time}}$ & \multicolumn{1}{r}{r\textsubscript{max}} & \multicolumn{3}{c}{gap values (\%)} & $\overline{\mbox{time}}$ \\
problem & (-) & (s) & (-) & r\textsubscript{max} & $\overline{\mbox{r}}$ ($\sigma$) & $\overline{\mbox{cost}}$ & (s) & problem & (-) & (s) & (-) & r\textsubscript{max} & $\overline{\mbox{r}}$ ($\sigma$) & $\overline{\mbox{cost}}$ & (s) \\
\hline
hex180   & 31.79 & 22  & 31.97 & 0.5  & -0.7 (0.4) & 0.2  & 111   & mesh115     & 37.71 & 9   & 38.57 & 2.3  & 0.7 (0.7) & 15.1  & 151   \\
hex240   & 28.90 & 21  & 29.03 & 0.5  & -0.2 (0.3) & 0.9  & 124   & mesh244     & 22.70 & 22  & 23.98 & 5.6  & 3.1 (1.4) & 16.5  & 994   \\
hex308   & 26.20 & 45  & 26.27 & 0.3  & -0.4 (0.3) & 1.0  & 281   & mesh268     & 20.66 & 25  & 23.08 & 11.7 & 9.0 (1.9) & 24.0  & 1996  \\
hex336   & 24.63 & 52  & 24.56 & -0.3 & -0.7 (0.2) & 0.4  & 316   & mesh293     & 18.29 & 27  & 21.03 & 14.9 & 11.8 (1.5) & 17.3  & 1727  \\
hex416   & 22.50 & 49  & 22.50 & 0.0  & -0.7 (0.3) & 0.5  & 366   & mesh343     & 18.58 & 41  & 20.30 & 9.3  & 5.5 (2.4) & 25.3  & 3553  \\
hex448   & 20.15 & 74  & 20.09 & -0.3 & -0.9 (0.3) & 0.7  & 524   & mesh374     & 18.64 & 31  & 19.28 & 3.4  & 1.3 (1.0) & 19.7  & 4126  \\
hex540   & 19.82 & 104 & 19.82 & 0.0  & -0.6 (0.4) & 0.0  & 473   & mesh400     & 16.64 & 42  & 17.61 & 5.8  & 2.7 (1.9) & 12.7  & 3175  \\
hex576   & 18.24 & 69  & 18.15 & -0.5 & -0.8 (0.3) & 0.0  & 295   & mesh449     & 16.34 & 41  & 17.38 & 6.3  & 4.8 (1.0) & 19.8  & 6431  \\
hex836   & 15.52 & 134 & 15.58 & 0.4  & -0.4 (0.5) & 0.0  & 625   & mesh686     & 11.89 & 70  & 13.56 & 14.0 & 10.1 (2.0) & 18.9  & 22760 \\
hex1144  & 13.39 & 227 & 13.39 & 0.0  & -0.6 (0.4) & 0.0  & 935   & mesh1337    & 8.44  & 197 & 9.28  & 9.9  & 5.6 (1.9) & 19.5  & 42390 \\
\hline
average  &       &     &       & 0.1  & -0.6 (0.3) & 0.4  &       & average     &       &     &       & 8.3  & 5.5 (1.6) & 18.9  &       \\
\hline
\hline
hexN180  & 24.52 & 12  & 25.29 & 3.1  & 1.5 (0.9) & 20.5  & 453   & sqr117  & 23.68 & 11  & 24.93 & 5.3  & 1.6 (1.3) & 9.9  & 133   \\
hexN240  & 24.18 & 20  & 25.11 & 3.8  & 1.9 (1.1) & 23.1  & 793   & sqr225  & 21.86 & 22  & 22.64 & 3.6  & 1.7 (0.9) & 16.4  & 743   \\
hexN306  & 23.34 & 27  & 24.73 & 6.0  & 4.8 (0.8) & 22.5  & 2148  & sqr260  & 22.08 & 27  & 22.94 & 3.9  & 1.6 (1.2) & 16.8  & 1564  \\
hexN336  & 23.79 & 33  & 24.55 & 3.2  & 1.6 (1.1) & 24.3  & 3209  & sqr297  & 21.86 & 31  & 22.50 & 3.0  & 0.6 (1.1) & 16.8  & 2186  \\
hexN416  & 23.45 & 47  & 24.50 & 4.5  & 2.2 (1.7) & 26.1  & 5929  & sqr341  & 21.74 & 31  & 22.56 & 3.8  & 1.5 (1.0) & 16.1  & 3093  \\
hexN448  & 23.60 & 44  & 24.60 & 4.3  & 2.1 (1.6) & 26.4  & 8503  & sqr377  & 21.29 & 43  & 22.08 & 3.7  & 2.4 (1.0) & 17.7  & 4843  \\
hexN510  & 23.51 & 54  & 24.12 & 2.6  & 0.9 (1.3) & 19.4  & 11991 & sqr400  & 21.40 & 51  & 21.94 & 2.5  & 1.3 (0.7) & 16.0  & 5469  \\
hexN576  & 23.76 & 68  & 24.30 & 2.3  & 0.2 (1.3) & 24.2  & 12357 & sqr448  & 21.27 & 60  & 21.95 & 3.2  & 0.8 (0.9) & 17.1  & 7047  \\
hexN798  & 23.09 & 104 & 23.97 & 3.9  & 1.5 (1.6) & 34.5  & 19531 & sqr676  & 21.03 & 105 & 21.47 & 2.1  & 0.5 (0.8) & 25.7  & 10864 \\
hexN1056 & 22.80 & 154 & 23.65 & 3.7  & 1.8 (1.5) & 38.9  & 28084 & sqr1015 & 20.78 & 190 & 21.11 & 1.6  & 0.1 (0.7) & 36.7  & 18130 \\
\hline
average  &       &     &       & 3.7  & 1.8 (1.3) & 26.0  &       & average &       &     &       & 3.3  & 1.2 (1.0) & 18.9  &       \\
\hline
\end{tabular}
\end{table*}

\begin{table}[ht]
\caption{TSP-SD results (fixed time of $10|V|$ sec.)}
\label{tab:GRASP}
\centering
\begin{tabular}{lrr|r|rr}
\hline
\multicolumn{3}{r|}{reference solver~\cite{Carmesin23}} & \multicolumn{3}{c}{proposed GRASP solver} \\
\multicolumn{2}{r}{time budget} & BKS & \multicolumn{1}{r}{cost\textsubscript{min}} & \multicolumn{2}{c}{gap values (\%)}   \\
problem     & (s)   & (-)                               & (-)                     & cost\textsubscript{min} & $\overline{\mbox{cost}}$ ($\sigma$)  \\
\hline
burma14     & 140   & 52     & 52     & 0.0   & 0.0  (0.0) \\
ulysses22   & 220   & 141    & 141    & 0.0   & 0.0  (0.0) \\
berlin52-10 & 520   & 23866  & 24046  & 0.8   & 1.9  (0.7) \\
berlin52-13 & 520   & 17263  & 17287  & 0.1   & 5.7  (2.2) \\
eil101      & 1010  & 1394   & 1323   & -5.1  & -1.9 (2.5) \\
gr202       & 2020  & 812    & 814    & 0.3   & 2.5  (0.9) \\
lin318      & 3180  & 110698 & 100706 & -9.0  & -4.6 (1.4) \\
fl417       & 4170  & 27162  & 22807  & -16.0 & -5.4 (3.8) \\
d657        & 6570  & 85054  & 79925  & -6.0  & -3.3 (1.3) \\
rat783      & 7830  & 13753  & 12427  & -9.6  & -6.7 (1.0) \\
vm1084      & 10840 & 325218 & 295995 & -9.0  & -6.3 (1.7) \\
\hline
average     &       &        &        & -4.9  & -1.7 (1.4) \\
\hline
\end{tabular}
\end{table}

\begin{table}[ht]
\caption{DTSP-CP results}
\label{tab:DTSP-CP}
\centering
\begin{tabular}{l|r|rrr|r}
\hline
                             & \multicolumn{1}{r}{r\textsubscript{max}} & \multicolumn{3}{c}{gap values (\%)}                                                                                    & $\overline{\mbox{time}}$ \\
problem-r\textsubscript{dub} & (-) & r\textsubscript{max} & $\overline{\mbox{r}}$ ($\sigma$) & $\overline{\mbox{cost}}$ ($\sigma$) & (s) \\
\hline
hex180-10                   & 31.10 & -2.7 & -3.8   (0.5)  & 0.0   (0.0)  & 478  \\
hex180-20                   & 31.05 & -2.9 & -3.7   (0.8)  & 0.0   (0.0)  & 533  \\
hex180-30                   & 30.67 & -4.1 & -4.9   (0.6)  & 0.5   (2.3)  & 800  \\
hexN180-10                  & 24.59 & -2.8 & -4.7   (2.0)  & 2.1   (3.6)  & 1085 \\
hexN180-20                  & 24.58 & -2.8 & -5.0   (3.2)  & 0.0   (0.0)  & 1014 \\
hexN180-30                  & 24.31 & -3.9 & -14.0  (11.3) & 21.1  (11.1) & 1927 \\
mesh115-10                   & 37.76 & -2.1 & -5.5   (2.1)  & 9.4   (8.3)  & 803  \\
mesh115-20                   & 37.06 & -3.9 & -6.5   (2.7)  & 0.7   (2.2)  & 814  \\
mesh115-30                   & 36.94 & -4.2 & -6.5   (1.6)  & 0.0   (0.0)  & 906  \\
sqr117-10                & 23.86 & -4.3 & -6.0   (1.2)  & 0.3   (1.3)  & 591  \\
sqr117-20                & 24.44 & -2.0 & -5.5   (1.9)  & 0.0   (0.0)  & 299  \\
sqr117-30                & 23.05 & -7.5 & -17.7  (12.6) & 21.1  (11.2) & 1434 \\
\hline
average                      &       & -3.6 & -7.0  (3.4)  & 4.6   (3.3)  &      \\
\hline
\end{tabular}
\end{table}

\subsection{TSP-SD experiments}

The performance of the proposed GRASP solver for the TSP-SD is documented in Tab.~\ref{tab:GRASP}.
The operators \emph{swap} and \emph{move} require parameters $k, l, m$.
Their values were selected from ranges $k,l,m \in \{1,2,3,4\}$ using the irace package~\cite{LopDubPerStuBir2016}.
The tuning resulted in the following setup: $m \in \{1, 2, 3\}$ and $(k, l) \in \{(1, 1), (1, 2), (2, 2), (2, 3), (3, 3)\}$.
Thus, including 2-opt, the GRASP solver uses 9 different operators in all the following experiments.

The TSP-SD was originally formulated in~\cite{Carmesin23} and solved with a generic metaheuristic solver~\cite{woller2022}.
This generic solver was tested on an artificial dataset of 11 instances, based on problems from the TSPLIB library~\cite{reinelt1991}, and the stop condition was given by a fixed computation time of $10|V|$ seconds, where $|V|$ is the number of nodes, indicated in each instance name.
E.g., the instance eil101 has $|V| = 101$ nodes and thus was solved for 1010 seconds.
The best-known solutions (BKS) from~\cite{Carmesin23} are used as a reference here.
The experimental setup is replicated, so the GRASP solver is tested with the same time budget of $10|V|$ seconds on the same machine.
Each instance is solved 50 times.
In Tab.~\ref{tab:GRASP}, we present gap values of $cost_{min}$ (best cost achieved over 50 runs), or $\overline{cost}$ (average cost over 50 runs) relative to the BKS of the reference solver~\cite{Carmesin23}.
The TSP-SD is a minimization problem, so a negative gap means better than the BKS result.


The two smallest instances with 14 and 22 nodes were solved to optimality in each run.
For the remaining 9 instances, the GRASP solver surpassed the BKS in 6 cases, both in the best and in the average performance.
The BKS was not improved for three instances with up to 202 nodes, but the GRASP solver still found a solution within the $0.8\%$ gap.
As for the 5 largest instances (318 to 1084 nodes), the BKS were substantially improved by $-9\%$ to $-16\%$.
The $\overline{cost}$ gaps range from $-3.3\%$ to $-6.7\%$ with an average standard deviation of 1.4\%, which indicates that the new solver is not only capable of improving the BKS, but robustly converges to better local optima on the large instances than the originally used generic solver.

\subsection{TSP-CP experiments}

The proposed TSP-CP solver is evaluated on four different datasets in Tab.~\ref{tab:TSP-CP_1}.
The datasets in the top row (hexagons, meshes) are adopted from~\cite{Kulich23}. 
The nodes in the hexagons instances are placed on a regular hexagonal grid.
The nodes in the mesh instances are randomly generated so that a certain minimal distance between them is maintained. 
The datasets in the bottom row (noisy hexagons, squares) are newly generated for this paper.
The noisy hexagons instances are similar to the hexagons instances, but the node positions are randomly shifted, so the grid is not perfectly regular.
Finally, the nodes in the square instances are placed on a regular square grid.
As a reference, the Best-Known Solutions (BKS) generated using the bisection algorithm for the WPCCP from~\cite{Kulich23} are presented in both tables, together with $\overline{time}$ (the average computation time).
As the TSP-CP is a relaxation of the WPCCP, this experiment primarily documents the effect of the relaxation on the maximal radius of the placed circles.
In Tab.~\ref{tab:TSP-CP_1}, we present the gap values of $r_{max}$ (best radius found by the proposed TSP-CP solver) and $\overline{r}$ (average radius over 20 runs) relative to BKS (the largest radius found) of the reference WPCCP solver~\cite{Kulich23}.
We also provide the gaps for the secondary objective $\overline{cost}$ (average cost of the TSP-CP cycle) relative to $cost_{TSP}$ (optimal solution cost of the corresponding TSP instance).
Note that $gap(r)$ is positive if $r$ is better than BKS, but $gap(cost)$ better than reference would be negative.
Each instance was solved 20 times.


Relaxing the WPCCP to the TSP-CP paid off most in the highly irregular meshes dataset, where the $gap(r_{max})$ is $8.3\%$ and in some mesh instances more than $10\%$.
The effect is lower in the noisy hexagons dataset ($3.7\%$) and regular squares dataset ($3.3\%$).
Interestingly, the new formulation and solver did bring only a negligible benefit of $0.01\%$ on the regular hexagons dataset.
In the WPCCP, the circles are placed around a fixed TSP cycle of minimal cost, whereas in the TSP-CP, the cycle is allowed to change in order to create additional space for placing larger circles.
The hexagonal structure appears to have some special property, which does not allow to take advantage of this TSP-CP feature.
The average computation times of the proposed solver are up to two orders higher for those instances, that can be improved by the new formulation. 
This significant difference in computation requirements is caused by the fact that the reference solver is used as a subroutine~\cite{Kulich23} in the proposed solver. 
Thus, the reference solver on its own is not capable of achieving the same scores no matter the computation time.
Regarding the cost of the TSP-CP solution, the gap from the optimal unconstrained TSP cycle $gap(\overline{cost})$ is often greater than $20\%$, except for the regular hexagons dataset.
It appears that a higher radius of placed circles means a higher cost, which is an intuitive conclusion.

\subsection{DTSP-CP experiments}

The DTSP-CP solver is tested on a subset of four instances, one from each dataset and the results are presented in Tab.~\ref{tab:DTSP-CP}.
Each instance is solved with three different Dubins turning radii $r_{dub} \in \{10, 20, 30\}$.
We provide the gap values of $r_{max}$ and $\overline{\mbox{r}}$ relative to $r_{max}^{TSP-CP}$ (largest radius obtained by the proposed TSP-CP solver on a given instance, Tab.~\ref{tab:TSP-CP_1}).
We also present gap of $\overline{cost}$ relative to $cost_{DTSP}$ (cost of the corresponding Dubins TSP solution).
The Dubins maneuvers are generated using the Dubins.jl library~\cite{DubLib}.

The average $gap(r_{max})$ over all instances and Dubins radii is $-3.6\%$, suggesting that the Euclidean TSP-CP variant generally allows placing larger circles.
The average $gap({\overline{cost}})$ is relatively lower than in the Euclid variant, which corresponds to the lower $gap(\overline{r})$.
In terms of computation times, the Dubins variant is around 4 to 8 times more expensive than the Euclid variant.
There are two possible reasons for this.
First, approximately $k^2|V|^2$ Dubins maneuvers must be calculated when solving the initial DTSP problem (line~\ref{alg:TSPCP_TSP}, Alg.~\ref{alg:TSPCP}).
Second, each time a TSP-SD instance is generated (line~\ref{alg:TSPCPf_TSPSDgen}, Alg.~\ref{alg:TSP-CP_fixed_r}), all edges $E$ in $G$ must be checked for collisions with newly placed circles $\mathcal{K}$.
Naturally, both of these operations are several times slower for the Dubins variant.

\section{Conclusions}

We present the Traveling Salesperson Problem with Circle Placement and its Dubins vehicle variant. Our contribution includes a novel heuristic solver for both variants, along with two solvers addressing related subproblems. We conduct a comparative analysis against state-of-the-art methods. In the future, we intend to modify the approach for placing circles of different radii aiming to maximize the area coverage.
Furthermore, we aim for an extension to account for multiple vehicles as well as including different types of motions. We also intend to investigate the theoretical bounds on the radii of the placed circles.

\bibliographystyle{IEEEtran}
\bibliography{IEEEabrv,sources}

\end{document}